\documentclass[10pt,twocolumn,letterpaper]{article}

\usepackage{wacv}
\usepackage{times}
\usepackage{epsfig}
\usepackage{graphicx}
\usepackage{amsmath}
\usepackage{amssymb}
\usepackage{booktabs}
\usepackage{multirow}
\usepackage[toc,page]{appendix}

\def\wacvPaperID{14} 

\wacvapplicationstrack 

\wacvfinalcopy 

\ifwacvfinal
\usepackage[breaklinks=true,bookmarks=false]{hyperref}
\else
\usepackage[pagebackref=true,breaklinks=true,colorlinks,bookmarks=false]{hyperref}
\fi

\begin{document}

\title{Multimodal Data Augmentation for Visual-Infrared Person ReID\\ with Corrupted Data}

\author{Arthur Josi, Mahdi Alehdaghi, Rafael M. O. Cruz, and Eric Granger\\
{\tt\small \{arthur.josi.1, mahdi.alehdaghi.1\}@ens.etsmtl.ca,
\{rafael.menelau-cruz, eric.granger\}@etsmtl.ca}
\and
Laboratoire d’imagerie, de vision et d’intelligence artificielle (LIVIA)\\
Dept. of Systems Engineering, ETS Montreal, Canada\\
}

\maketitle
\thispagestyle{empty}

\begin{abstract}
The re-identification (ReID) of individuals over a complex network of cameras is a challenging task, especially under real-world surveillance conditions. Several deep learning models  have been proposed for visible-infrared (V-I) person ReID to recognize individuals from images captured using RGB and IR cameras. However, performance may decline considerably if RGB and IR images captured at test time are corrupted (e.g., noise, blur, and weather conditions). Although various data augmentation (DA) methods have been explored to improve the generalization capacity, these are not adapted for V-I person ReID.  
In this paper, a specialized DA strategy is proposed to address this multimodal setting. Given both the V and I  modalities, this strategy allows to diminish the impact of corruption on the accuracy of deep person ReID models. Corruption may be modality-specific, and an additional modality often provides complementary information. Our multimodal DA strategy is designed specifically to encourage modality collaboration and reinforce generalization capability. For instance, punctual masking of modalities forces the model to select the informative modality. Local DA is also explored for advanced selection of features within and among modalities.
The impact of training baseline fusion models for V-I person ReID using the proposed multimodal DA strategy is assessed on corrupted versions of the SYSU-MM01, RegDB, and ThermalWORLD datasets in terms of complexity and efficiency. Results indicate that using our strategy provides V-I ReID models the ability to exploit both shared and individual modality knowledge so they can outperform models trained with no or unimodal DA. GitHub code: \url{https://github.com/art2611/ML-MDA}.
\end{abstract}

\section{Introduction}

Real-world monitoring and surveillance application (e.g., individuals in airport, and vehicles in traffic) rely on challenging tasks, like object detection \cite{zou2019object, zaidi2022survey}, tracking \cite{luo2021multiple}, and re-identification (ReID) \cite{khan2019survey, PersonREID_outlook}. The aim of person ReID is to recognize individuals over a set of distributed non-overlapping cameras. State-of-art systems for person re-identification (e.g., deep Siamese networks) typically learn an embedding through various metric learning losses, which aim at making similar image pairs (with the same identity) closer to each other and dissimilar image pairs (with different identities) more distant from each other. Despite the recent advances with deep learning (DL) models, person ReID remains a challenging task due to the non-rigid structure of the human body, the different viewpoints/poses with which a person can be observed, image corruption, and the variability of capture conditions (e.g., illumination, scale, contrast) \cite{bhuiyan2020pose, mekhazni2020unsupervised}. 
    
Visible-infrared (V-I) person ReID aims to recognize individuals of interest across a network of RGB and IR cameras. IR cameras are often employed in conjunction with RGB cameras for, e.g., night time recognition in outdoor environments. Most approaches for V-I person ReID focus on the cross-modal matching problem. This paper focuses on person ReID systems that allow for fusion of visible and infrared modalities based on a joint representation space. Although several techniques have been proposed for dynamic and attention-based fusion \cite{ismail2020improving,msaf2020su}, few V-I person ReID methods have been proposed for RGB-IR fusion \cite{RegDB}. In this setting, it is difficult to extract discriminant modality-specific features when one modality becomes corrupted, while conserving the shared modality features \cite{baltruvsaitis2018multimodal}. 
    
In real-world surveillance applications, the accuracy of person ReID models often declines when image data is corrupted by noise, occlusions, saturation, blur, weather conditions, etc. \cite{chen2021benchmarksDataset_C}. Several strategies have been developed to improve the generalization performance of person ReID models in response to corrupted image data. Using more complex DL models, trained with more data have been shown to improve the performances in object detection \cite{michaelis2019benchmarking}, and image classification \cite{xie2020self} tasks. For instance, using transformer-based models may be more suitable to tackle corruption \cite{hendrycks2020pretrained, chen2021benchmarksDataset_C}. However, using more complex models, like vision-transformers \cite{han2020survey} limits real-time ReID applications.  In addition, using more diverse training data can help \cite{xie2020self}, and therefore data augmentation (DA) \cite{chen2021benchmarksDataset_C} methods may improve performance, without increasing the models complexity, and while avoiding the costs of data collection and annotation \cite{shorten2019survey}. 

In this paper, we propose a MDA strategy to improve the accuracy of V-I person ReID systems.
Chen \etal \cite{chen2021benchmarksDataset_C} recently proposed a DA learning strategy, called the Consistent ID Loss, with Inference before BNNeck, and Local-based Augmentation (CIL). It is mainly based on local DA, and provides improvements in accuracy for unimodal (RGB) person ReID. However, the multimodal aspect has not been explored in the literature to tackle corruptions. Yet, such approach might be helpful to tackle corruption as modalities are not similarly affected by corruptions and can still benefit by DA strategies \cite{hao2022mixgen}. 
    
To manage corrupted image data in multimodal settings, a multimodal DA (MDA) strategy is introduced, allowing to leverage the complementary knowledge among modalities, while dynamically balancing the importance of individual modality in the final predictions. Consequently, the strategy should reduce the corruption impact. Having in mind the multimodal person ReID aspect, and regarding that person ReID datasets were only used for cross-modal ReID, protocols are provided along with a comprehensive study over three V-I person ReID datasets, SYSU-MM01 \cite{SYSU}, RegDB \cite{RegDB} and (less explored)  ThermalWORLD \cite{ThermalWORLD}. Finally, as the focus is made on corruption robustness for the multimodal setting, the corruption benchmark proposed by \cite{chen2021benchmarksDataset_C} is extended to the infrared thermal modality.

Our main contributions are summarized as follows. 
(1) A MDA strategy is proposed to improve the accuracy of DL models for V-I person ReID. To optimize the collaboration among modalities, discriminant joint feature representations in the DL model, our MDA strategy relies on local occlusions and global modality masking data augmentation. 
(2) A comprehensive V-I multimodal experimental protocol is proposed to evaluate the impact on performance of clean and corrupted image data using the well-known SYSU-MM01, RegDB, and ThermalWORLD datasets. Corruptions from \cite{chen2021benchmarksDataset_C} are extended to the infrared domain to analyse multimodal data corruption impact.
(3) An extensive set of experiments is conducted, showing that the used V-I fusion model outperforms the related state-of-art models. The limitations of unimodal models are shown by comparing a basic fusion model learned with the adapted DA to the unimodal state-of-art person ReID models.

\section{Related Work}\label{sec:related_work}

\noindent \textbf{A) Multimodal person ReID.}
Most approaches for person ReID in the last decade \cite{PersonREID_outlook} focus on the unimodal (RGB) \cite{ristani2018features,luo2019bag} and cross-modal \cite{hao2021cross,alehdaghi2022visible,zhang2022fmcnet} settings. Few focused on combining multimodal information, despite the potential to improve performance in the joint representation setting \cite{baltruvsaitis2018multimodal}. For example, Chen \etal extracted contours from the RGB modality and used a two-stream CNN architecture to combine information \cite{chen2019contour}. Bhuiyan \etal proposed to use pose information to gate the flow of visual information through a CNN backbone \cite{bhuiyan2020pose}. These approaches used the knowledge extracted from the main modality, which would be similarly affected by image corruption. 

Some approaches sought to leverage the complementarity of RGB and depth modalities for an accurate person ReID \cite{paolanti2018person,lejbolle2018attention,martini2020open}. However, Nguyen \etal \cite{RegDB} represents the only approach where visible and infrared modalities are integrated into a joint representation space. Infrared and visual features are concatenated from embeddings extracted independently trained CNNs, and used for pairwise matching at test time. This simple model attained an impressive performances on the RegDB dataset. However, RegDB data is captured with only one camera per modality, and RGB-IR cameras are co-located, with only a single tracklet of ten images per modality and individual. For these reasons, the RegDB dataset is less consistent with a real-world scenario. In fact, the development of person ReID models that are effective in uncontrolled real-world scenario remains an open problem  \cite{hendrycks2021many}. 

\noindent \textbf{B) Corruption and data augmentation strategies.}    
Data augmentation (DA) consists in multiplying the available training dataset by punctually applying transformations on training images, like flips, rotations, and scaling \cite{ciregan2012multi}. This way, a model usually benefits from increased robustness to image variations, and improved generalization performance. According to Geirhos \etal \cite{geirhos2018generalisation}, training a model on a given corruption is not often helpful over other types of degradation. Yet, \cite{rusak2020simple} showed that a well-tuned DA can help the model to perform well over multiple types of image corruption, through Gaussian and Speckle noise augmentation. Hendrycks \etal proposed the Augmix strategy \cite{hendrycks2019augmix}, where various transformations are randomly applied to an image, and then mix multiple of those augmented images. Random Erasing punctually occludes parts of the images by replacing pixels with random values \cite{zhong2020random}. Those strategies allow a large variety of augmented image, simulating eventually real-world data, and hence inducing higher generalization performance. 

Focusing on person ReID, Chen \etal \cite{chen2021benchmarksDataset_C} proposed both a corrupted RGB dataset (adapted from \cite{hendrycks2019benchmarking}) and the CIL learning strategy to improve systems performance under corrupted data. Their strategy is partly based on two local DA methods -- self-patch mixing and soft random erasing. The former replaces some of the pixels in a patch with random values, while the latter superposes a randomly selected patch from an image at a random position on this same image. Gong \etal \cite{gong2021person} show interesting improvements through local and global grayscale patch DA on RGB images. The previous strategies are limited to single modality stream models, even though the latter shows how greyscale data may reinforce the visible modality features using DA. MDA strategies have presented encouraging results for image-text emotion recognition \cite{xu2020mda} or vision-language representation learning \cite{hao2022mixgen}. However, to our best knowledge, our work is first to propose MDA with V-I person ReID applications. 
    \begin{figure}[t]
        \begin{center}
        \includegraphics[width=0.99\linewidth]{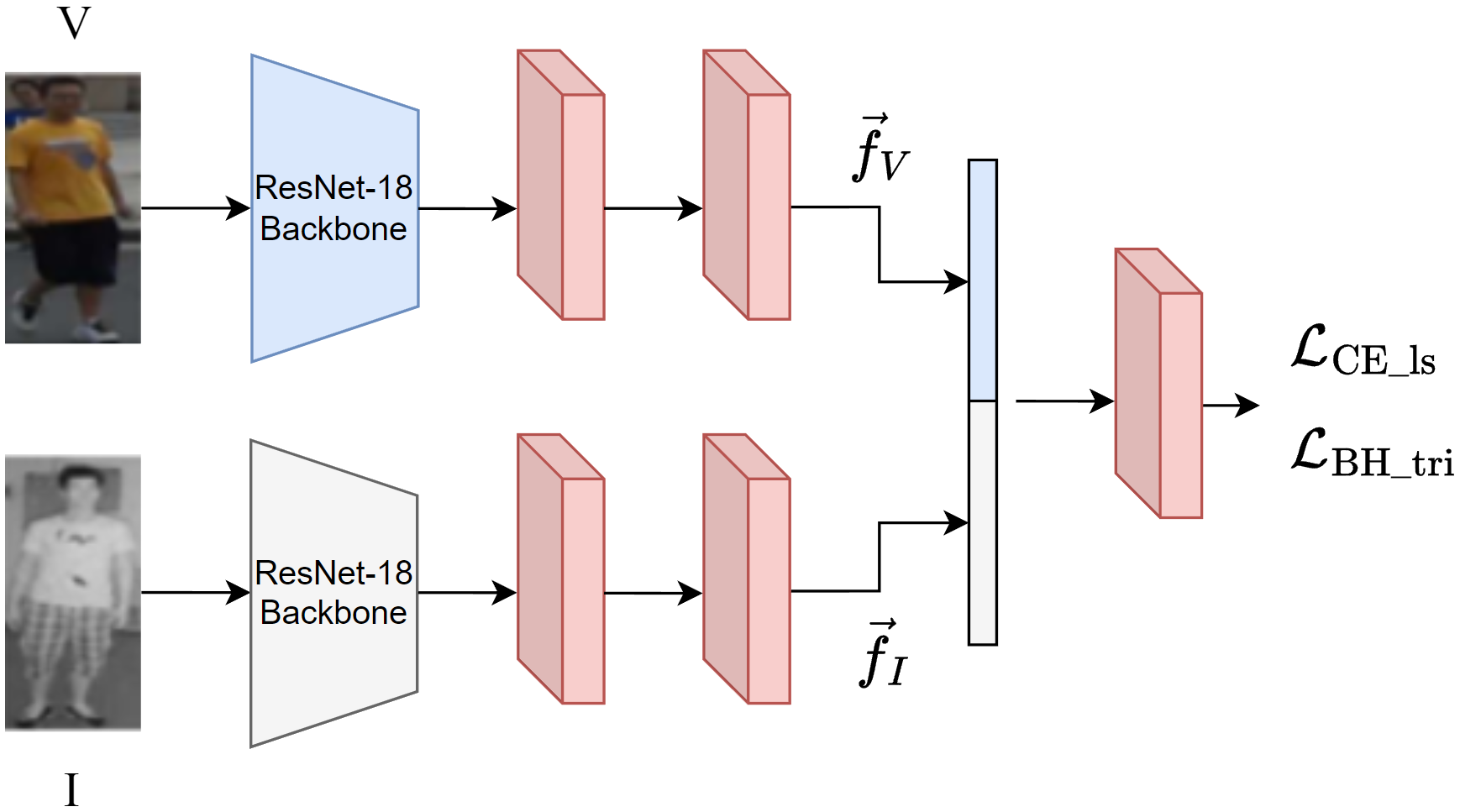}
        \end{center}
        \caption{Training architecture considered for V-I person ReID. It learns a joint multimodal representation by concatenating features produced by independent I and V ResNet-18 CNN backbones.
        }
        \label{fig:Model_multimodal}
    \end{figure}

\section{Proposed Strategies}
    
    \begin{figure*}
    \begin{center}
    \includegraphics[width=1.00\linewidth]{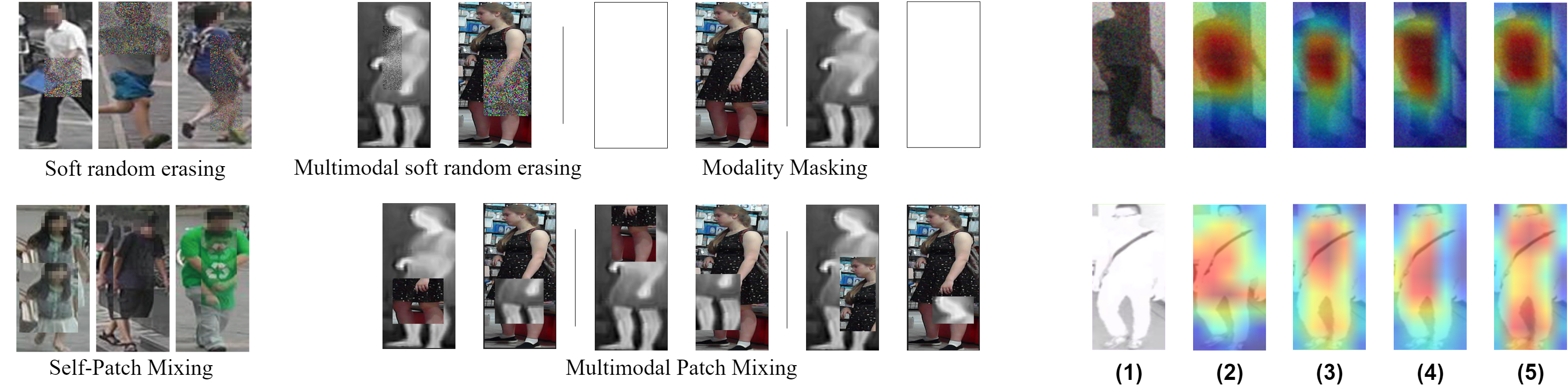}
    \end{center}
      \caption{\textbf{Left} present data augmentation methods from \cite{chen2021benchmarksDataset_C}. \textbf{Center} are our augmentation methods extensions from those, along with the proposed Modality Masking approach. \textbf{Right} shows visualizations of activation maps from corrupted a sample and from differently trained models. \textbf{(1)} Pair input data, RGB corrupted through Gaussian noise and IR through saturation. \textbf{(2)} Augmix. \textbf{(3)} Multimodal Patch Mixing. \textbf{(4)} Modality Masking. \textbf{(5)} Multimodal soft random erasing. The discriminability increases from left to right.} 
    \label{fig:data_augmentation_strategies}
    \end{figure*}
    
Our strategy is based on co-learning, allowing each modality stream to adapt to the other \cite{baltruvsaitis2018multimodal}. Using our MDA strategy, we expect to adapt DL models from one modality stream to another, and consequently provide better robustness to corrupted multimodal data. 
The low-cost multimodal architecture that we considered for V-I person ReID is based on two parallel ResNet-18 \cite{resnet_he2016deep} backbones pre-trained on ImageNet \cite{Imagenet}. Rather than having a large single stream model, such architecture might allow us to present a competitive model both in size and efficiency. After the two backbones, each stream has an average pooling and a batch normalization layer. The final prediction is obtained by  concatenating features from each embedding, right before presenting it to a fully connected layer (Fig. \ref{fig:Model_multimodal}). Embeddings are concatenated during the test phase for pairwise similarity matching, from which the final ranking is obtained. 
    
\noindent \textbf{A) Multimodal patch mixing and soft random-erasing.} Making a multimodal model focus on modality-specific features is challenging, as the model usually mainly focuses on shared features \cite{baltruvsaitis2018multimodal}. Augmenting data with local occlusions may help the model to emphasize modality-specific feature importance, as some features will be available only from one or the other modality. 

Multimodal soft Random Erasing (MS-REA): The soft random erasing (S-REA) \cite{chen2021benchmarksDataset_C} might play this role, as it occludes parts of the RGB image punctually, potentially letting the opportunity for the hetero modality to close this occlusion gap. For S-REA, a proportion of the pixels in a given patch are given random values. To make the model close the occlusion gap in a bi-directional manner, the MS-REA is proposed (Fig. \ref{fig:data_augmentation_strategies}), applying grayscaled random pixel values on a given path of the thermal modality, as well as the random values pixel values on the RGB modality. Grayscale values respect the infrared thermal image definition as IR thermal is encoded on one channel, potentially aligning better with real-world corruptions.  
    
Multimodal Patch Mixing (M-PATCH): Our M-PATCH DA inspired by the Self Patch (S-PATCH) DA \cite{chen2021benchmarksDataset_C}. Through M-PATCH, the idea is to extract a patch from each modality and superimpose it on the hetero-modality. The IR modality receives the RGB patch from the same individual, and vice versa. As the patches come from the same individual, the model has the option to rely on the patch features to discriminate. Three variants are explored which have different disturbance levels. From the less disturbing to the most disturbing, the first variation is extracting the patch from the Same part of the image, and applying it at the Same location on both modalities (-SS). The second extracts from the Same location but apply at Different locations (-SD), and the third extracts from Different locations and also apply at Different locations (-DD) (Fig. \ref{fig:data_augmentation_strategies}). The M-PATCH approach might gather the best of both RandomPatch \cite{zhou2019omnirandompatch} and S-PATCH \cite{chen2021benchmarksDataset_C} strategies. RandomPatch is strongly disturbing, and the model is forced to focus on out-of-patch features as the patch gathers information related to a different individual. S-PATCH less disturbing -- it allows the model to focus on in-patch features as it contains features related to the same individual. Ours also allows in-patch feature selection by using the same individual, but provides more disturbance since the patch comes from the hetero modality. This approach may reinforce the model's shared features finding, while also pushing the model to exchange information across modalities.

\noindent \textbf{B) Modality masking.}
A modality might be punctually unavailable or primarily uninformative. Though, the model has to know how to cancel a modality so that this one should not have a high impact in the final prediction. The modality masking approach is expected to make the model learns such behavior, by punctually replacing one or another modality with an entirely blank image. Instead of masking the multimodal representation as it has been done in \cite{gabeur2022masking}, a representation is extracted from the masked input, so the model has to learn how to cancel its influence on the final results. This masking DA is expected to complement the previously presented DA. The M-PATCH and MS-REA approaches supposedly focus on making the model better at selecting the right features within a modality. The idea is here to balance the importance of each modality in the final embedding regarding the level of corruption of each.

\section{Results and Discussion}

\subsection{Datasets and performance measures}

    Since our study is focused on V-I multimodal person ReID, we employ the widely known SYSU-MM01 (SYSU) \cite{SYSU} and RegDB \cite{RegDB} datasets, along with the lesser-known ThermalWORLD (TWORLD) \cite{ThermalWORLD} dataset. Details on these datasets are show in Table \ref{tab:datasets_details}), allowing us to evaluate under diverse conditions. 
    \begin{table}[htbp]
      \centering
      \small
      \caption{Statistics of SYSU, RegDB, and TWORLD datasets. \textbf{V}: \textbf{V}isible and \textbf{I}:  \textbf{I}nfrared. Image size and number per identity is presented as: Min;Max;Avg. BRISQUE \cite{mittal2011blindBRISQUE} metric as: avg±std.}
        \begin{tabular}{l||c|c|c}
         \midrule
        & \textbf{SYSU} & \textbf{RegDB} & \textbf{TWORLD} \\
        \midrule
        V-images  & 29,033 & 4,120  & 8,125 \\
        I-images  & 15,712 & 4,120  & 8,125 \\
        V-Camera  & 4     & 1     & 16 \\
        I-Camera  & 2     & 1     & Generated \\
        Identities  & 491   & 412   & 409 \\
        Paired cameras & No    & Yes   & Yes \\
        V-images/id & 10;144;59.1 & 10;10;10& 1;155;19.9 \\
        I-images/id & 10;144;32.0 & 10;10;10 & 1;155;19.9 \\
        Image width & 26;1198;111 & 64;64;64 & 10;810;141  \\
        Image height & 65;879;291 & 128;128;128 & 25;897;353 \\   
        V-BRISQUE & 30.50±12.26 & 38.84±9.86 & 27.79±13.28 \\
        I-BRISQUE & 40.52±8.42 & 38.81±9.56 & 60.25±8.67 \\
        \midrule
        \end{tabular}%
      \label{tab:datasets_details}%
    \end{table}%

    \noindent \textbf{SYSU-MM01.} \cite{SYSU} gather 4 RGB and 2 thermal cameras, with 491 distinct individuals, $29 033$ RGB, and $15 712$ IR thermal images. The specificity of this dataset is that its RGB and IR cameras are not co-located. 
    
    \noindent \textbf{RegDB.} \cite{RegDB} is a much smaller dataset, with one camera only per modality, co-located cameras, and a single 10 images tracklet per identity and camera. RegDB $410$ identities lead to $4 120$ images per modality.  
    
    \noindent \textbf{ThermalWORLD.} \cite{ThermalWORLD} is only partially available, leading us to $409$ distinct identities and $8125$ RGB images from 16 cameras. IR images were generated synthetically. Hence, cameras can be considered as co-located. However, the thermal images are of poor quality (see BRISQUE \cite{mittal2011blindBRISQUE} value of $60.25$ in Table \ref{tab:datasets_details}). 
    
    \noindent \textbf{Corruptions.} For comparison reasons, the corruptions used by Chen \etal \cite{chen2021benchmarksDataset_C} are the same in this study. However, the RGB corruptions were adapted to the thermal modality (detailed in supplementary material) as the thermal modality would more likely get impacted in a real scenario. The RGB data corruptions proposed by \cite{chen2021benchmarksDataset_C} are mentioned through the notation \textbf{-C}, and its extension with both modalities corrupted through the notation \textbf{-C*}. Corruptions are applied independently and randomly for the RGB and the IR modalities and on both the query and the gallery images to match real-world conditions.
    
    \noindent \textbf{Performance Measures.} The mean Average Precision (mAP), and the mean Inverse Penalty (mINP) are used as performance metrics, commonly used for person ReID \cite{PersonREID_outlook}.

\subsection{Implementation details}\label{sec:implementationDetails}
    
    \noindent \textbf{Data division.} SYSU-MM01 and RegDB datasets have well-established V-I cross-modal protocols \cite{Join_Pixel_align, wang2019rgb, wang2019learning, ye2019bi}, but multimodal protocols remain to be built. Following SYSU-MM01 authors' cross-modal protocol, 395 identities were used for the training set, and 96 identities were used for the testing set. For RegDB, the 412 identities are kept as well into the two identical sets of 206 individuals. The SYSU-MM01 train/test ratio is kept for ThermalWORLD, leading to 325 training identities and 84 for testing. A 5-fold validation \cite{K-cross-validation} is performed over the data used for training, using folds of respectively $79$, $41$, and $65$ distinct identities for SYSU-MM01, RegDB, and ThermalWORLD.
    
    \noindent \textbf{Data augmentation (DA).} The Augmix, S-PATCH, or S-REA were evaluated following the original papers settings. Our proposed multimodal extensions M-PATCH and MS-REA were used with the same appearance augmentation probability as S-PATCH and S-REA. Modality Masking is applied randomly on one or another modality, with equiprobability, and occurs with a default probability of $1/8$. 
    For RegDB, the validation set uses the same DA as the training set. This way, better performances were observed, since they maxed out in the early epochs, or otherwise do not learn complex cues for the model. 
    
    \noindent \textbf{Pre-processing.} A data normalization is done at first by re-scaling RBG and IR images to $144 \times 288$. Random cropping with zero padding and horizontal flips are adopted for base DA. Those parameters were proposed by \cite{PersonREID_outlook} on RegDB and SYSU-MM01 datasets. The same normalization is kept under ThermalWORLD for consistency among protocols. 

    \begin{table*}
      \centering
      \caption{The performance of various multimodal DA strategies using a standard model (V-I ReID model trained without DA) as baseline. Augmix DA is applied with and without other proposed DA approaches.}
        \begin{tabular}{l||cc|cc|cc|cc|cc|cc}
        \midrule
              & \multicolumn{2}{c|}{\textbf{SYSU}} & \multicolumn{2}{c|}{\textbf{SYSU-C*}} & \multicolumn{2}{c|}{\textbf{RegDB}} & \multicolumn{2}{c|}{\textbf{RegDB-C*}} & \multicolumn{2}{c|}{\textbf{TWORLD}} & \multicolumn{2}{c}{\textbf{TWORLD-C*}} \\
              \textbf{DA Strategy} & mAP   & mINP  & mAP   & mINP  & mAP   & mINP  & mAP   & mINP  & mAP   & mINP  & mAP   & mINP \\
        \midrule
        Standard & 96.47 & 73.69 & 25.01 & 1.90  & 99.64 & 98.46 & 21.80 & 2.40  & 87.90 & 49.05 & 29.30 & 3.93 \\
        Augmix \cite{hendrycks2019augmix} & 95.37 & 68.60 & 35.23 & 2.56  & 99.88 & 99.40 & 40.75 & 9.10  & 87.12 & 46.33 & 42.26 & 5.69 \\
        \midrule
        + S-REA \cite{chen2021benchmarksDataset_C}& 96.21 & 74.36 & 43.24 & 4.06  & 99.90 & 99.51 & 43.84 & 10.25 & 89.24 & 50.10 & 54.14 & 8.92 \\
        + MS-REA & \textbf{96.81} & \textbf{77.02} & \textbf{61.44} & \textbf{8.34}  & 99.86 & 99.35 & \textbf{57.84} & \textbf{19.38} & 88.95 & 49.92 & \textbf{58.10} & \textbf{9.89} \\
        \midrule
        + S-PATCH \cite{chen2021benchmarksDataset_C} & 96.40 & 74.89 & 31.39 & 2.14  & 99.90 & 99.53 & 41.83 & 9.39  & 89.12 & 50.53 & 40.73 & 5.63 \\
        + MS-PATCH & 94.70 & 69.10 & 33.69 & 2.17  & 99.89 & 99.41 & 40.97 & 9.34  & \textbf{89.26} & \textbf{51.26} & 41.75 & 5.57 \\
        + M-PATCH-SS & 96.10 & 73.40 & 35.49 & 2.44  & 99.86 & 99.34 & 43.28 & 10.68 & 88.35 & 50.16 & 44.41 & 5.61 \\
        + M-PATCH-SD & 95.94 & 72.93 & 35.10 & 2.40  & 99.87 & 99.35 & 42.95 & 10.31 & 88.58 & 51.59 & 43.49 & 5.53 \\
        + M-PATCH-DD & 94.98 & 68.95 & 33.90 & 2.42  & 99.89 & 99.48 & 41.98 & 9.71 & 88.49 & 51.35 & 43.90 & 5.51 \\
        \midrule
        + Masking & 95.61 & 73.49 & 40.92 & 2.90  & \textbf{99.90} & \textbf{99.52} & 49.27 & 12.10 & 86.01 & 42.76 & 39.91 & 6.16 \\
        \midrule
        \end{tabular}%
      \label{tab:ablation_study}%
    \end{table*}%
    
    \noindent \textbf{Hyperparameters.} The hyperparameters values in our models were set based on the default AGW \cite{PersonREID_outlook} baseline. The SGD is used for training optimization, combined with a Nesterov momentum of $0.9$, and a weight decay of $5e-4$. Our models are trained through $100$ epochs. Early stopping is applied based on validation mAP performances. The learning rate is initialized at $0.1$ and follows a warming-up strategy \cite{warming_up_strat}. The batch size is set to 32, with 8 distinct individuals and 4 images per individual. The paired image is selected by default for RegDB and ThermalWORLD. For the SYSU-MM01 dataset, the images from the hetero modality are randomly selected through the available ones for a given identity. 
    
    \noindent \textbf{Losses.} The Batch Hard triplet loss \cite{TripletLoss} $\mathcal{L}_{\text{BH\_tri}}$ and the cross-entropy with regularization via Label smoothing \cite{szegedy2016rethinking} $\mathcal{L}_{\text{CE\_ls}}$ are used as loss functions for our models. Indeed, the former is widely used in person ReID approaches \cite{wang2019rgb,Hi-cmd,PersonREID_outlook}, so the same margin value is fixed at $0.3$, and the latter is part of the CIL implementation \cite{chen2021benchmarksDataset_C}. The total loss corresponds to the sum of both losses. The batch hard triplet loss aims at reducing the distance in the embedding space for the hardest positives while increasing the distance for the hardest negatives. The regularization with label smoothing works at reducing the gap between logits, which makes the model less confident on predictions and hence improves generalization \cite{muller2019does}.
    
    \noindent \textbf{Leave-one-out query strategy.} \label{LOO} The single-shot and the multi-shot settings \cite{wang2019rgb} are widely used in cross-modal papers to form the query and gallery sets. For these settings, one or ten images from the hetero modality are selected per identity and camera to join the gallery, while the other modality forms the probe set. However, such an approach is not so realistic in a surveillance context, as the video makes the gallery number of frames per person vary much. These variations cannot be controlled as individuals are unknown in the final environment. Hence, a new strategy is developed, inspired by the leave-one-out cross-validation strategy \cite{K-cross-validation}, named Leave-One-Out Query (LOOQ). The LOOQ strategy treats the extreme but meaningful case in which one would have only a unique image of the person to ReID and multiple footages containing images of this same person in the gallery. Every pair of images is alternatively used as a probe set while all the other pairs join the gallery. This allows us to respect the original dataset statistics (see Table \ref{tab:datasets_details}) by authorising the gallery images per individual to vary. Also, the mINP metric relates to the hardest test sample from the same individual. Hence, computing this metric over multiple gallery images makes it more consistent, appearing even more important in a corrupted context.
    
    Concerning the implementation, the images are paired for both RegDB and ThermalWORLD datasets, so the paired image from the hetero modality joins the query and gallery set directly during the formation of those sets. However, SYSU-MM01 needs personal treatment since its images are not paired. Plus, the image number per modality for a given individual varies (Table \ref{tab:datasets_details}). To solve this issue, as many pairs of images as possible are randomly selected with the constraint that one image from one modality or another must not appear in two distinct pairs. Because random image pairs are formed for SYSU-MM01, a mean of 30 trials is performed to present robust and reliable results according to the Central Limit Theorem. 
   
\subsection{Benchmarking data augmentation strategies} \label{sec:benchmark_data_aug}
    
    Table \ref{tab:ablation_study} shows the impact on person ReID performance of each DA strategy is investigated over the three datasets under clean and corrupted (-C*) settings. First, we compare the model learned without DA (Standard) with the model learned with Augmix, and the models learned with Augmix plus other augmentation. The other DA strategies can be S-REA, S-PATCH, or one of our proposed augmentation.  
    
    \noindent \textbf{Multimodal soft random erasing.} The S-REA strategy applies random values to a certain proportion of the pixels in a given patch of the RGB image. A good improvement can be seen from the Augmix to the S-REA strategy for each dataset and the clean and corrupted settings. Still, a more significant improvement happened for ThermalWORLD-C* compared to SYSU-MM01-C* and RegDB-C*, respectively, with a $11.88$\% improvement against $8.01$\% and $3.09$\%. While extending the DA to the multimodal setting through MS-REA, we observe a remarkable improvement for each corrupted setting, and especially that the improvement is much higher on both SYSU-MM01 and RegDB compared to ThermalWORLD. Indeed, mAP increases by $18.20$\% and $14.00$\% for SYSU-MM01-C* and RegDB-C* respectively against $3.96$\% for ThermalWORLD-C*. ThermalWORLD has a much weaker IR modality, so the model probably focuses much on the visible modality. Consequently, the model probably almost fully benefits from S-REA as if it were a unimodal architecture. The other datasets do not allow to benefit as much from this DA, as the model has presumably learned to focus more on IR due to the unbalanced augmentation (applied only on RGB). In contrast, the equilibrium brought by MS-REA probably allows the full exploitation of the approach and explains the impressive improvement from S-REA to MS-REA. Also, MS-REA comes first among approaches under the clean setting for SYSU-MM01 and RegDB datasets, except for ThermalWORLD. With a $95$\% confidence, results using MS-REA compared to the best approach are not statistically significant for RegDB, whereas it is for ThermalWORLD according to the Cochran p-values \cite{K-cross-validation} of respectively $0.29$ and $4.89e-5$. Thanks to MS-REA and partial occlusions, the model might have learned not to only focus on the most discriminant cues, as confirmed by the IR activation map comparison from Augmix to MS-REA (see Fig. \ref{fig:data_augmentation_strategies}). Also, this approach present important improvement over biased data augmentation, denoting a great generalization power (detailed in supplementary material).

    \begin{table*}[htbp]
      \centering
      \caption{Data augmentation combination. Each is used along with Augmix and MS-REA. C1 stands for Masking, C2 for M-PATCH-SS and C3 for Masking and M-PATCH-SS.}
        \begin{tabular}{l||cc|cc|cc|cc|cc|cc}
         \midrule
               & \multicolumn{2}{c|}{\textbf{SYSU}} & \multicolumn{2}{c|}{\textbf{SYSU-C*}} & \multicolumn{2}{c|}{\textbf{RegDB}} & \multicolumn{2}{c|}{\textbf{RegDB-C*}} & \multicolumn{2}{c|}{\textbf{TWORLD}} & \multicolumn{2}{c}{\textbf{TWORLD-C*}} \\
              \textbf{Strategy} & mAP   & mINP  & mAP   & mINP  & mAP   & mINP  & mAP   & mINP  & mAP   & mINP  & mAP   & mINP \\
        \midrule
        MS-REA & \textbf{96,81} & \textbf{77,02} & 61,44 & 8,34  & 99.86 & 99.35 & 57.84 & 19.3  & 88,95 & 49,92 & 58,10 & 9,89 \\
        MS-REA + C1 & 96.77 & 76.01 & 63.01 & 9.59  & 99.90 & 99.45 & \textbf{61.92} & \textbf{20.14} & 86.34 & 43.24 & 56.10 & 11.04 \\
        MS-REA + C2 & 96.85 & 75.87 & 61.19 & 9.13  & 99.85 & 99.26 & 56.23 & 17.98 & \textbf{89.16} & \textbf{50.68} & 57.45 & 9.64 \\
        MS-REA + C3 & 96.78 & 75.87 & \textbf{63.83} & \textbf{9.77} & \textbf{99.89} & \textbf{99.48} & 61.53 & 20.17 & 86.65 & 43.75 & \textbf{57.95} & \textbf{11.53} \\
         \midrule
        \end{tabular}%
      \label{tab:combination}%
    \end{table*}%
    
    \noindent \textbf{Multimodal patch mixing.} Observing the results obtained for SYSU-MM01 and ThermalWORLD, the performances globally improved from the Augmix strategy to the S-PATCH approach for the clean datasets, while those are reduced under the corrupted setting. While applying the self patch mixing on both modalities through MS-PATCH, performances are questionable, as performances remains lower or equivalent to Augmix on corrupted data, while conserving or decreasing from clean S-PATCH results. In practice, it is only while considering the modality patch exchange in our M-PATCH strategy, especially the less disturbing version M-PATCH-SS, that the best improvement is obtained on the corrupted setting, while conserving great performances on the clean one. Indeed, mAP is respectively improved by $2.15$\% and $2.53$\% over the Augmix strategy for ThermalWORLD-C* and RegDB-C*. The cameras might need to be co-located for the approach to perform, as SYSU-MM01-C* pretty much conserve similar performances as Augmix on corrupted data, and as the standard model on clean data. Spatial alignment is probably helping much the model to find correlations between the hetero modality patch and the current modality image. Still, there is a performance improvement on two datasets even if this one remains much lower than the previous MS-REA approach. 
    
    \noindent \textbf{Masking.} The modality masking approach presents interesting improvements under the SYSU-MM01 and RegDB datasets. Indeed, performances on corrupted datasets are increased by $5.69$\% mAP and $8.52$\% mAP over the Augmix approach, while those are pretty much matching Augmix performances on the clean datasets. The modality masking DA consists of punctually feeding a modality stream with a fully uninformative modality. Hence, those results show that the approach can make the model better able to give more importance to the discriminant modality for a given pair of images. The ability to balance modality influence on not co-located cameras through SYSU-MM01 dataset is important to highlight. Concerning the ThermalWORLD dataset, the Masking model's performances decrease for the clean and the corrupted setting compared to Augmix. Indeed, the mAP is respectively lower by $1.11$\% and $2.35$\%. Such a decrease is not surprising, as this dataset's thermal modality is very uninformative. Hence, while learning, a masked visible modality probably acts as noise by creating not discriminant V-I pairs. 
    
    \noindent \textbf{Combination.} As the DA approaches have distinct expected roles in the way they help the model to get more robust against corruption, combining them might allow to benefit from each of their specificities (Table \ref{tab:combination}). It is interesting to see that the real combining improvement comes from the Masking approach used with MS-REA (C1) on both SYSU-MM01-C* and RegDB-C*, with respectively $1.57$\% and $4.08$\% improvement over MS-REA used by itself. ThermalWORLD did not benefit from the masking DA, which could be expected as the Masking was already decreasing its performance when used alone. Adding M-PATCH to MS-REA (C2) or to MS-REA and Masking (C3) seems not to bring meaningful additional improvements. Indeed, MS-REA + (C3) matches the performances of MS-REA + (C1) under the clean and corrupted settings on both RegDB and SYSU-MM01. Similar observations can be done from MS-REA alone and MS-REA + M-PATCH. Hence, even if M-PATCH has shown improvements on RegDB and ThermalWORLD when used alone, those improvements are probably mainly due to the benefits of occlusions, which are already part of the MS-REA approach. Visual results observing especially IR activation maps seem to confirm this aspect (Fig. \ref{fig:data_augmentation_strategies}). Though, using MS-REA with M-PATCH appear as not being meaningful.
    From the previous conclusions, we propose the Masking and Local Multimodal Data Augmentation (ML-MDA) strategy, which combines both the local approach MS-REA with the modality masking DA.

    \begin{table*}[htbp]
      \centering
      \small
      \caption{Performance of our multimodal model using ML-MDA compared against SOTA unimodal person ReID models, and a ResNet-18 unimodal model while using CIL or not. The two last rows show the performance of the same model when RGB is corrupted (-C), and when RGB and IR are corrupted (-C*). Note that performance on clean datasets are the same and presented in fused cells.}
    
        \begin{tabular}{l||cc|cc|cc|cc|cc|cc}
        \midrule
        \textbf{Model}      & \multicolumn{2}{c|}{\textbf{SYSU}} & \multicolumn{2}{c|}{\textbf{SYSU-C}} & \multicolumn{2}{c|}{\textbf{RegDB}} & \multicolumn{2}{c|}{\textbf{RegDB-C}} & \multicolumn{2}{c|}{\textbf{TWORLD}} & \multicolumn{2}{c}{\textbf{TWORLD-C}} \\
              & mAP   & mINP  & mAP   & mINP  & mAP   & mINP  & mAP   & mINP  & mAP   & mINP  & mAP   & mINP \\
        \midrule
        ResNet-18 & 86,25 & 39,97 & 32,36 & 1,91  & 99,26 & 96,64 & 45,15 & 5,68  & 86.44 & 49.44 & 28.06 & 3.86 \\
        TransReID [1] & 94,33 & 64,79 & 52,03 & 3,60  & 99,34 & 97,35 & 45,64 & 5,69  & \textbf{95.86} & \textbf{77.98} & 65.47 & 17.20 \\
        LightMBN [2] & 94,45 & 64,06 & 40,90 & 2,13  & 99,90 & 99,41 & 32,40 & 3,25  & 93.02 & 65.94 & 37.34 & 5.60 \\
        \midrule
        ResNet-18 + CIL & 86,64 & 42,78 & 51,64 & 3,83  & 99,65 & 98,41 & 55,76 & 10,98 & 86.95 & 48.07 & 52.85 & 7.97 \\
        TransReID [1] + CIL & 93,20 & 62,02 & 61,38 & 7,20  & 99,69 & 98,57 & 58,74 & 12,89 & 94.79 & 73.82 & \textbf{73.61} & \textbf{23.16} \\
        LightMBN [2] + CIL & 94,07 & 61,95 & 67,80 & 8,23  & 99,89 & 99,41 & 66,55 & 21,53 & 93.20 & 66.14 & 71.30 & 19.73 \\
        \midrule
        Ours + ML-MDA (-C) & \multirow{2}{*}{\textbf{96.77}} & \multirow{2}{*}{\textbf{76.01}}  & \textbf{87.89} & \textbf{42.5}  & \multirow{2}{*}{\textbf{99.90}}  & \multirow{2}{*}{\textbf{99.45}}  & \textbf{92.37} & \textbf{75.71} & \multirow{2}{*}{86.34}  & \multirow{2}{*}{43.24} & 69.20 & 18.47 \\
        Ours + ML-MDA (-C*) &  &  & 63.01 & 9.59  & & & 61.92 & 20.14 & & & 56.10 & 11.04 \\
        \midrule
     \end{tabular}%
    
      \label{tab:SOTA_comparison}%
    \end{table*}%
    
\subsection{Comparison with the state-of-art}\label{sec:SOTA}
 
    \noindent \textbf{Performance.} As there is no true competitor in the area of V-I multimodal person ReID, the ML-MDA strategy is compared with SOTA unimodal person ReID models, with or without the CIL strategy used. According to  results obtained in \cite{chen2021benchmarksDataset_C}, the LightMBN \cite{herzog2021lightmbn} and TransReID \cite{he2021transreid} models are respectively the most performing unimodal models under the clean and corrupted scenarios. For fair comparison, Table \ref{tab:SOTA_comparison} shows two scenarios for the multimodal test data. First, both RGB and IR are corrupted (-C*), and second, to observe how a clean IR modality can help when the RGB modality is corrupted, performance is also compared when only RGB is corrupted (-C).
    
   Considering a clean data setting, the ML-MDA model outperforms the second-best approach by $2.32$\% mAP and especially by $11.22$\% mINP on SYSU-MM01. This significant mINP improvement shows that the multimodal setting helps considerably on the more challenging images. Indeed, the multimodal model can compensate for challenging RGB samples with the IR modality. On the RegDB dataset, our approach outperforms the others, with a statistically significant improvement. Indeed, with a $95$\% confidence interval, the Cochran \cite{K-cross-validation} p-value between LightMBN, LightMBN + CIL and our approach is of $0.02$. On ThermalWORLD, the performance of the multimodal model cannot compare with TransReID and LightMBN models, not even improving over the ResNet-18 model. Again, the poor quality IR mostly acts as a source of perturbation for the model. 

    When the RGB modality only is corrupted (-C) on both SYSU-MM01 and RegDB datasets, the ML-MDA model provides a considerable performance improvement over TransReID and LightMBN models. Indeed, our model reaches $87.98$\% mAP and $92.37$\% mAP for SYSU-MM01 and RegDB, respectively improving by $20.09$\% and $25.82$\% over the second-best approach. These improvements highlight the benefits of a well-trained multimodal model, relying mainly on the clean modality (I) when the other is corrupted (V). A performance gap between -C to -C* settings can be observed, as -C* is much more challenging with two corrupted modalities. The multimodal model appears as the second-best approach for both SYSU-MM01-C* and RegDB-C* datasets. Indeed, LightMBN reaches respectively $67.80$\% and $66.55$\% mAP against $63.01$\% and $61.92$\% mAP for our multimodal model. Still, the multimodal setting improves the mINP for SYSU-MM01, from $8.23$\% to $9.59$\%, and is only below RegDB by $1.39$\% mINP, showing that the multimodal setting can help on the hardest cases. For the -C* setting, our approach outperforms other models except for LightMBN + CIL, or on ThermalWORLD data, apparently unable to encode discriminant cues from the corrupted IR to counterbalance the well designed unimodal models. However, our architecture remains very simple, and obtaining such performance improvement on our light architecture is already promising. More complex fusion strategies, with more knowledge exchange between modality streams, and a more robust backbones like ResNet-50, may allow exceeding the performance of LightMBN.
    
    Note that both the -C and the -C* settings might not be the most accurate for a great multimodal evaluation. Indeed, considering the IR always clean (-C) is not so accurate as weather would, for example, probably happen on both modalities for a given co-located pair. On the other hand, considering both modalities always corrupted (-C*) hardly allows the hetero modality to help the primary modality, but is not so realistic either. Indeed, digital corruption or noise would probably not affect V-I modalities simultaneously. In fact, the real-world setting would allow [Clean RGB, Corrupted IR] pairs, and would especially be a mixture of -C and -C* settings. In practice, there should be more pairs in which one of the two modalities remains clean, so the true potential of the multimodal setting probably lies somewhere in between the -C and the -C* settings.  

    \begin{table}[htbp]
      \centering
      \caption{Memory (number of parameters) and time (FLOPs) complexity of proposed and baseline ReID models, FLOPs computed from a single or multi-modal input.}
        \begin{tabular}{l||r|r}
        \midrule
        \textbf{Model} & \textbf{No. Params (M)} & \textbf{FLOPs (G)} \\
        \midrule
        ResNet-18                           & 11.3      & 0.51 \\
        TransReID \cite{he2021transreid}    & 102.0     & 19.55 \\
        LightMBN \cite{herzog2021lightmbn}  & 7.6       & 2.09 \\
        \midrule
        Ours                                & 22.5      & 1.54 \\
        \midrule
        \end{tabular}%
      \label{tab:models_complexity}%
    \end{table}%
    
    \noindent \textbf{Complexity.} Multimodal person ReID with IR and RGB is more complex than regular ReID with RGB, so models are compared in terms of number of parameters and FLOPs (Table \ref{tab:models_complexity}). The TransReID \cite{he2021transreid} model is known for being computationally expensive as its architecture is transformer based, with a total of 102M parameters and 19.55 G-FLOPs. In contrast, LightMBN \cite{herzog2021lightmbn} is based on the Os-Net architecture, which makes it very light, requires $7.6M$ parameters and $2.09G$ FLOPs. Even if our multimodal model has more parameters ($22.5M$) to adjust than LightMBN, it requires less memory compared to the SOTA unimodal person ReID models, with its $1.54G$ FLOPs. Although our model seems equivalent to LightMBN in terms of complexity, it provides a significant performance improvement. Its robustness to corrupted data makes it an excellent trade-off in the face of uncontrollable scenarios.

\section{Conclusion}
Real-world surveillance often requires light models that perform well on corrupted data. In this paper, image corruptions were extended to the infrared modality, and MDA strategy was proposed to improve the performance of the V-I person ReID. Experiments on the SYSU-MM01, RegDB and ThermalWORLD datasets showed the benefits of the multimodal setting over SOTA unimodal ReID models, especially when combined with the specialized MDA strategy. Indeed, our ML-MDA strategy has allowed for significant improvements in terms of robustness to corruption using the proposed modality masking and MS-REA MDA. The former learns the model to dynamically balance the importance of each modality in the final embedding. The latter works on the occlusion concept and teaches the model to better select features among modalities and not to focus only on the most discriminant features. ML-MDA improves performances, yet does not incur additional model complexity, and allows for a light ReID architecture. 

Given multiple modalities, MDA allows addressing image data corruption, as these corruptions impact V and I modalities in a different ways, allowing the hetero-modality to compensate. MDA could be studied more independently from person ReID, and our methods can be applied to more general datasets (e.g., RGB-D data). Moreover, increasing the number modalities could further reduce the impact of corruption.
Note that potential improvements are possible using more advanced fusion methods \cite{msaf2020su,ismail2020improving}. Finally, we believe that our multimodal corrupted test set might not entirely reflect the true potential of the multimodal setting, as discussed section \ref{sec:SOTA}. To better fit real-world conditions, corruption correlations among modalities should be considered in the test set design. This would probably allow the multimodal setting to perform even better. 

\vspace*{0.2cm}
\noindent \textbf{Acknowledgements:} This research was supported by Nuvoola AI Inc., and the Natural Sciences and Engineering Research Council of Canada.

{\small
\bibliographystyle{ieee_fullname}
\bibliography{egbib}
}

\begin{appendices}

    \section{Thermal modality corruptions:}\label{app:thermal_modality_corruptions}

\begin{figure*}
    
    \centering
    \includegraphics[width=0.6\textwidth]{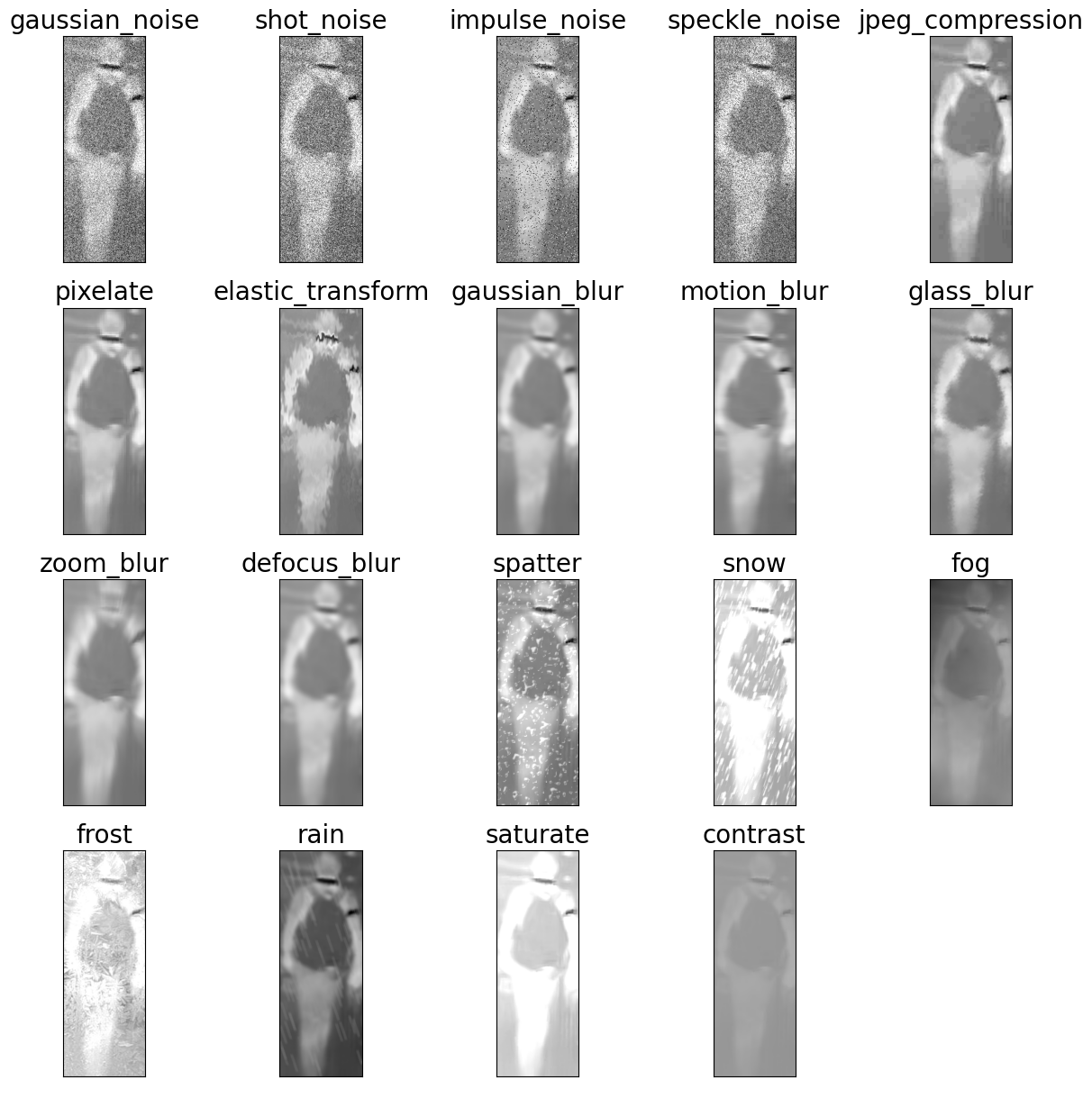}
    \caption[width=0.6\textwidth]{Taxonomy of our 19 used thermal corruptions, all applied on level 3.}
    \label{fig:thermal_corruptions_level3}
    
    \end{figure*}

    In practice, except for the brightness corruption, each and every other corruptions from the 20 corruptions \cite{chen2021benchmarksDataset_C} appeared to be meaningful for the thermal modality as well. Concerning the brightness changes, the thermal modality is not impacted by the current luminosity so this corruption is removed. Then, all others applicable, with eventually few slight adjustments that we describe here. Those corruptions are visible Fig. \ref{fig:thermal_corruptions_level3}. The different forms of noises, like Gaussian, Shot, Impulse, Speckle are applied similarly, except that we turned the noise into greyscale values while combining it with the original IR images. Speckle and frost are two other corruptions which needed to be grayscaled before being applied on the thermal images. Indeed, they initially apply eventual color changes on the images, with blue colored water or brown colored dirt for spatter, and blue frozen masks for frost. As a last adjustment, the saturation is expressed differently for the IR modality, brightening eventually the object of interest if this one is too close to the camera. Regarding this expression of the thermal saturation, we applied in practice the brightness function as ease to mimic it. Then, every other corruptions are affecting similarly both modalities, so we applied them the exact same ways.
    
    \section{Additional experiments - Intuitive data augmentation}\label{app:intuitive}
    
    For additional interpretations, two common corruptions that are part of the corrupted test sets are used as a data augmentation strategy. 
    
    \noindent \textbf{Approach.} As the idea is to keep a degree of blindness with the test set encountered corruptions, those two data augmentation corruptions are applied at a fixed intensity. The first data augmentation is the Gaussian blur corruption, applied with a blur radius of $3$ for both the RGB and the IR. The latter is the RGB's luminosity changes and the IR modality's image saturation. The luminosity changes are whether increasing RGB brightness by an enhancement factor $2$ or decreasing it by a factor $0.5$. The saturation for the IR is changed by increasing the enhancement factor to $1.5$. Each of those augmentation occurs with equiprobability on the RGB and the IR modality, with a probability of $1/8$. As those corruptions are part of the corrupted test set, which biased the obtained performances, these results will only be used for comparison but are not proposed as a solution for better generalization. 
    
    \noindent \textbf{Results.} The obtained results are gathered table \ref{tab:complete_ablation_study} Looking at the reached performances by blur data augmentation or luminosity and saturation ones, it appears that those strategies are significantly improving over Augmix on the -C* setting. From those results only, we cannot know how well it allows the model to generalize over other types of corruption. However, it is clear that it helps and probably comes from better handling the related corruptions in the test corrupted test set. Still, the Masking strategy compares well with those for both SYSU-MM01-C* and RegDB-C* while not introducing bias in the results, which is a great observation. Also, the MS-REA approach is much ahead of those results, showing that the strategy allows a great generalization power while not introducing such bias to the results. 
    
    \begin{table*}
  \centering
  \small
  \caption{Comparison of the various data augmentation strategies, with standard being the multimodal concatenation model without specific augmentation, to standard Augmix \cite{hendrycks2019augmix} and Augmix with one or another data augmentation. This table gather random activation, Patch, Masking and intuitive corruptions related strategies.}
    \begin{tabular}{l|cc|cc|cc|cc|cc|cc}
          & \multicolumn{2}{c|}{\textbf{SYSU}} & \multicolumn{2}{c|}{\textbf{SYSU-C*}} & \multicolumn{2}{c|}{\textbf{RegDB}} & \multicolumn{2}{c|}{\textbf{RegDB-C*}} & \multicolumn{2}{c|}{\textbf{Tworld}} & \multicolumn{2}{c}{\textbf{Tworld-C*}} \\
          & mAP   & mINP  & mAP   & mINP  & mAP   & mINP  & mAP   & mINP  & mAP   & mINP  & mAP   & mINP \\
    \midrule
    Standard & 96.47 & 73.69 & 25.01 & 1.90  & 99.64 & 98.46 & 21.80 & 2.40  & 87.90 & 49.05 & 29.30 & 3.93 \\
    Augmix \cite{hendrycks2019augmix} & 95.37 & 68.60 & 35.23 & 2.56  & 99.88 & 99.40 & 40.75 & 9.10  & 87.12 & 46.33 & 42.26 & 5.69 \\
    \midrule
    + S-REA \cite{chen2021benchmarksDataset_C}& 96.21 & 74.36 & 43.24 & 4.06  & 99.90 & 99.51 & 43.84 & 10.25 & 89.24 & 50.10 & 54.14 & 8.92 \\
    + MS-REA & \textbf{96.81} & \textbf{77.02} & \textbf{61.44} & \textbf{8.34}  & 99.86 & 99.35 & \textbf{57.84} & \textbf{19.38} & 88.95 & 49.92 & \textbf{58.10} & \textbf{9.89} \\
    \midrule
    + S-PATCH \cite{chen2021benchmarksDataset_C} & 96.40 & 74.89 & 31.39 & 2.14  & 99.90 & 99.53 & 41.83 & 9.39  & 89.12 & 50.53 & 40.73 & 5.63 \\
    + MS-PATCH & 94.70 & 69.10 & 33.69 & 2.17  & 99.89 & 99.41 & 40.97 & 9.34  & \textbf{89.26} & \textbf{51.26} & 41.75 & 5.57 \\
    + M-PATCH-SS & 96.10 & 73.40 & 35.49 & 2.44  & 99.86 & 99.34 & 43.28 & 10.68 & 88.35 & 50.16 & 44.41 & 5.61 \\
    + M-PATCH-SD & 95.94 & 72.93 & 35.10 & 2.40  & 99.87 & 99.35 & 42.95 & 10.31 & 88.58 & 51.59 & 43.49 & 5.53 \\
    + M-PATCH-DD & 94.98 & 68.95 & 33.90 & 2.42  & 99.89 & 99.48 & 41.98 & 9.71 & 88.49 & 51.35 & 43.90 & 5.51 \\
    \midrule
    + Masking & 95.61 & 73.49 & 40.92 & 2.90  & \textbf{99.90} & \textbf{99.52} & 49.27 & 12.10 & 86.01 & 42.76 & 39.91 & 6.16 \\
    \midrule
    + Blur & 94.77 & 69.38 & 41.72 & 3.06  & 99.86 & 99.33 & 45.68 & 11.66 & 88.05 & 51.36 & 50.36 & 7.66 \\
    + Lum - Sat  & 94.99 & 70.31 & 38.54 & 2.79  & 99.79 & 99.00 & 54.05 & 17.89 & 88.01 & 50.25 & 44.09 & 5.73 \\
    \end{tabular}%
  \label{tab:complete_ablation_study}%
\end{table*}%

\end{appendices}

\end{document}